\documentclass[conference]{IEEEtran}
\IEEEoverridecommandlockouts
\usepackage{cite}
\usepackage{amsmath,amssymb,amsfonts}
\usepackage{algorithmic}
\usepackage{graphicx}
\usepackage{textcomp}
\usepackage{subcaption}
\usepackage{xcolor}
\usepackage{multirow,tabularx}
\usepackage{booktabs}
\usepackage[normalem]{ulem}
\def\BibTeX{{\rm B\kern-.05em{\sc i\kern-.025em b}\kern-.08em
    T\kern-.1667em\lower.7ex\hbox{E}\kern-.125emX}}
\makeatletter

\begin{document}

\title{Learned Hierarchical B-frame Coding with Adaptive Feature Modulation for YUV 4:2:0 Content}

\author{
    \begin{tabular}{cccccc}
        Mu-Jung Chen$^{1}$ &
        Hong-Sheng Xie$^{1}$ &
        Cheng Chien$^{1}$ &
        Wen-Hsiao Peng$^{1}$ &
        Hsueh-Ming Hang$^{2}$
    \end{tabular}\\
    
    $^1$Computer Science Dept., 
    $^2$Electronics Engineering Dept., 
     National Yang Ming Chiao Tung University, Taiwan
    \vspace{-2ex}
}


\maketitle

\begin{abstract}
This paper introduces a learned hierarchical B-frame coding scheme in response to the Grand Challenge on Neural Network-based Video Coding at ISCAS 2023. We address specifically three issues, including (1) B-frame coding, (2) YUV 4:2:0 coding, and (3) content-adaptive variable-rate coding with only one single model. Most learned video codecs operate internally in the RGB domain for P-frame coding. B-frame coding for YUV 4:2:0 content is largely under-explored. In addition, while there have been prior works on variable-rate coding with conditional convolution, most of them fail to consider the content information. We build our scheme on conditional augmented normalized flows (CANF). It features conditional motion and inter-frame codecs for efficient B-frame coding. To cope with YUV 4:2:0 content, two conditional inter-frame codecs are used to process the Y and UV components separately, with the coding of the UV components conditioned additionally on the Y component. Moreover, we introduce adaptive feature modulation in every convolutional layer, taking into account both the content information and the coding levels of B-frames to achieve content-adaptive variable-rate coding. Experimental results show that our model outperforms x265 and the winner of last year's challenge on commonly used datasets in terms of PSNR-YUV.

\end{abstract}

\vspace{-0.5em}
\section{Introduction}



Despite the fact that the recent developments of end-to-end learned video compression have shown promising coding performance, there remain many issues to be addressed. Most of the existing works~\cite{dvclu,ssf,nvc,fvc,mlvc,hlvc,rafc,elfvc,rlvc,accv} focus on P-frame coding, while B-frame coding, which allows the use of both the future and past reference frames for higher coding efficiency, is largely under-explored. Among the others, B-frame coding needs to address excessive motion overhead and the efficient use of the two reference frames. Besides, most learned codecs operate internally in the RGB 4:4:4 domain, even when the input is YUV 4:2:0 content. The conversation from YUV 4:2:0 to RGB 4:4:4 is introduced before compression to tackle the uneven spatial resolutions of the Y (luminance) and UV (chrominance) components. It is also common that multiple networks are used to achieve variable-rate compression, which is impractical in many real-world applications.

There have been few attempts at B-frame coding. Wu~\emph{et al.}~\cite{Wu_2018} perform contextual coding of a B-frame based on the motion-compensated, multi-scale features extracted from the two reference frames. The idea is coined conditional (inter-frame) coding in~\cite{iclrw}, which additionally introduces one-stage, conditional motion coding without estimating flow maps explicitly. Following a more traditional approach, the works in~\cite{Djelouah, hlvc, murat_lhbdc} first encode two optical flow maps derived from the past and future references, followed by synthesizing a predicted frame for residual coding in the feature or pixel domain. To reduce motion overhead, Pourreza~\emph{et al.}~\cite{BEPIC} interpolate a predicted frame as a reference frame for coding a B-frame with the P-frame codec.


To achieve YUV 4:2:0 coding, Egilmez~\emph{et al.}~\cite{egilmez2021transform} propose a branched network that processes Y and UV components separately before fusing their latents into a combined representation for coding. The other straightforward approaches include applying the space-to-depth conversion of the Y component, or coding Y and UV components separately. The latter calls for separate Y and UV codecs. These ideas are studied for image compression only. 

To achieve variable-rate compression, it is common to introduce conditional convolution to adapt feature distributions~\cite{cconv, gvae, saft} according to a rate-dependent hyperparameter. Lately, the idea of conditional convolution is extended to video coding~\cite{icip21} and~\cite{iscas22}, to achieve variable-rate compression with a single model.





In response to the Grand Challenge on Neural Network-based Video Coding at ISCAS 2023~\cite{iscas_gc23}, we propose a hierarchical B-frame coding system for YUV 4:2:0 content. Inspired by~\cite{canfvc}, we adopt conditional augmented normalizing flows (CANF) to perform conditional motion and inter-frame coding. In particular, we encode the Y and UV components by two separate conditional codecs, where the coding of the UV components is conditioned additionally on the Y component. Moreover, to achieve content-adaptive variable-rate coding with a single model, \textcolor{black}{we extend conditional convolution to accommodate not only the rate parameter, but also the coding levels of B-frames and the output features of the convolutional layers. Experimental results show that our model outperforms x265~\cite{x265} and the winner~\cite{iscas22} of last year's challenge~\cite{iscas_gc} on commonly used datasets in terms of PSNR-YUV.}




\section{Related Work}

\begin{figure}[t!]
    \begin{subfigure}{\linewidth}
        \centering
        \includegraphics[width=0.9\linewidth]{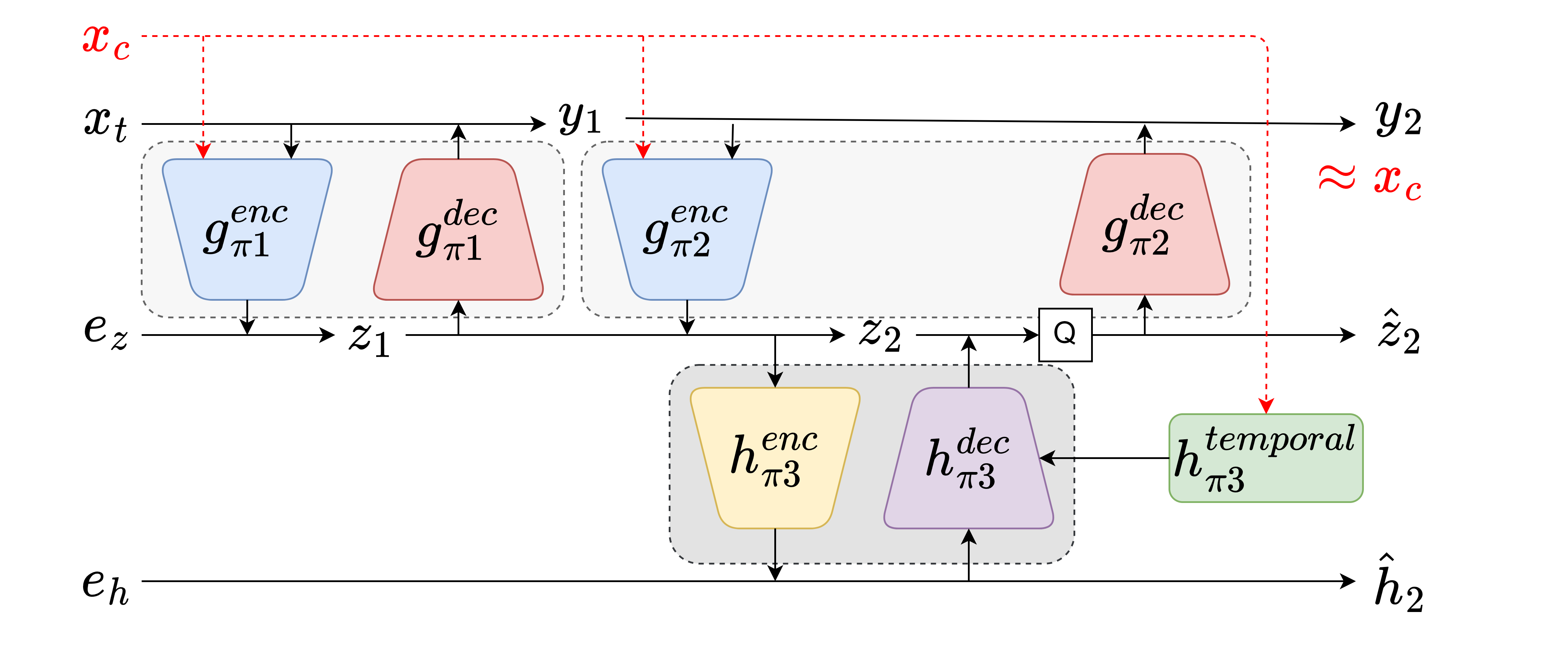}
        \vspace{-1em}
        \caption{}
    \end{subfigure}
    
    \begin{subfigure}{\linewidth}
        \centering
        \includegraphics[width=0.9\linewidth]{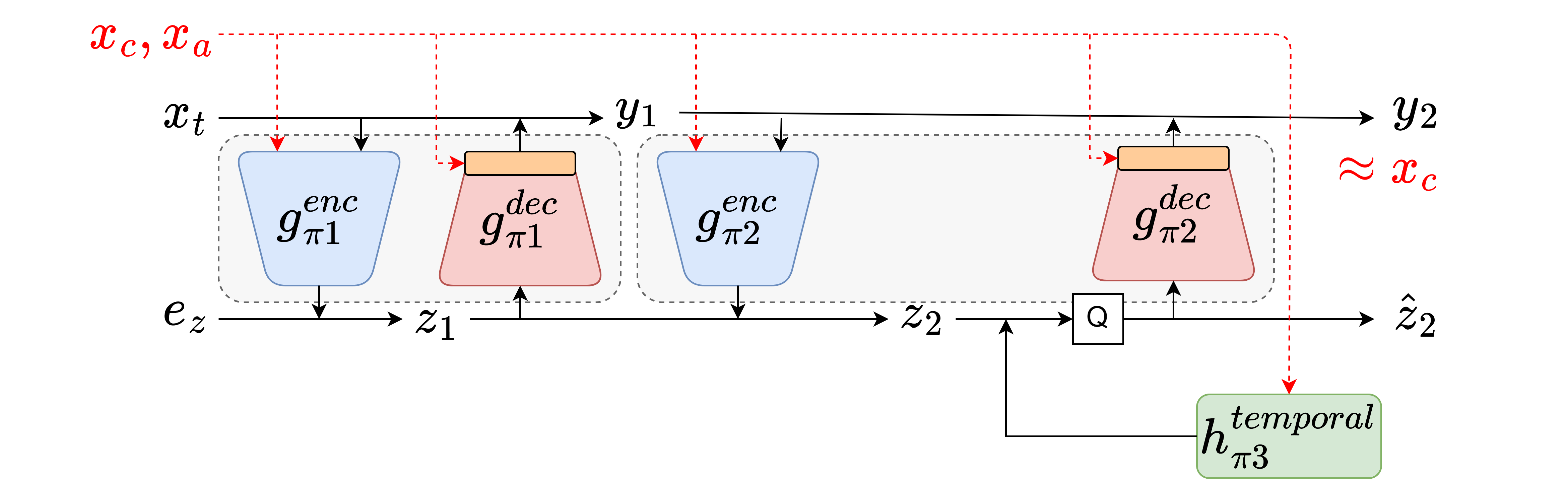}
        \vspace{-0.5em}
        \caption{}
    \end{subfigure}
    \caption{\textcolor{black}{Illustration of (a) the CANF-based inter-frame codec, which is used for coding the Y component in this work, and (b) the CANF-based inter-frame codec for coding the UV components, where $x_a$ represents the coded Y component.}}
    \label{fig:canfvc}
    \vspace{-1em}
\end{figure}

This section reviews the basics of the CANF-based inter-frame coding~\cite{canfvc}, to ease the understanding of our scheme. Fig.~\ref{fig:canfvc} illustrates its architecture for inter-frame coding, which aims to learn the conditional distribution $p(x_t|x_c)$ of the coding frame $x_t$ given its motion-compensated reference frame $x_c$. In~\cite{canfvc}, this is achieved by maximizing the conditional augmented likelihood $p(x_t,e_z,e_h|x_c)$, where $e_z, e_h$ are the two augmented noise inputs. From the operational perspective, $x_c$ serves as the conditioning factor in the autoencoding transforms, composed of $\{g_{\pi_i}^{enc},g_{\pi_i}^{dec}\}_{i=1}^2$, which convert (from left to right) $x_t$ into $x_c$, with the quantized latent $\hat z_2$ capturing the information needed to instruct the conversion and $\hat{h}_2$ taking the role of the hyperprior. The conversion between $x_t$ and $x_c$ is approximate and lossy. The same CANF-based codec can also be utilized to encode optical flow maps conditionally. To this end, a flow map predictor must be created to serve as the condition.      



\vspace{-0.5em}
\section{Proposed Method}

\subsection{System Overview}
\label{sec:system_overview}
Fig.~\ref{fig:arch} presents an overview of our proposed method. 
As shown, the encoding of a B-frame $x^{420}_t$ begins with using the motion estimation network (MENet) operating internally in the YUV 4:4:4 domain to obtain bi-directional optical flow maps $m_{t\to t-k}, m_{t\to t+k}$ according to its two reference frames $\hat{x}^{420}_{t-k}, \hat{x}^{420}_{t+k}$, respectively. The resulting flow maps are compressed jointly by the CANF-based conditional motion codec ($M, M^{-1}$) given the conditioning signals $m^p_{t\to t-k}, m^p_{t\to t+k}$ generated by the motion prediction network (MPNet). The decoded flow maps $\hat{m}_{t\to t-k}, \hat{m}_{t\to t+k}$ are used for bi-directional motion compensation. Particularly, we adopt two separate motion compensation networks (MCNet-\textit{Y}, MCNet-\textit{UV}) to synthesize the motion-compensated frames $\hat{x}^y_c, \hat{x}^{uv}_c$ for Y and UV components, respectively. $\hat{x}^y_c, \hat{x}^{uv}_c$ serve as the conditioning signals for conditional inter-frame coding of $x^y_t, x^{uv}_t$ to obtain the reconstructed Y and UV components $\hat{x}^y_t, \hat{x}^{uv}_t$, respectively. Notably, for coding the UV components, we introduce the reconstructed Y component as an additional conditioning signal. The following sections elaborate on these proposed modules.

\subsection{Conditional Bi-directional Motion Coding}
\label{sec:motion_coding}
The proposed conditional motion codec follows the same CANF-based design as Fig.~\ref{fig:canfvc} (a). In the present context, we concatenate $m_{t\to t-k}, m_{t\to t+k}$ to form a 4-channel input for coding (i.e. $x_t$ in Fig.~\ref{fig:canfvc} (a) becomes the concatenated signal from $m_{t\to t-k}, m_{t\to t+k}$). Likewise, the predicted flow maps $m^p_{t\to t-k}, m^p_{t\to t+k}$ output by the motion prediction network (MPNet) are concatenated to replace $x_c$ in Fig.~\ref{fig:canfvc} (a) as the conditioning signal for motion coding. This allows the motion codec to exploit freely the correlation inherent in $m_{t\to t-k}, m_{t\to t+k},m^p_{t\to t-k},~\text{and}~m^p_{t\to t+k}$.


\subsection{Motion Estimation and Compensation Networks}
\label{sec:motion_compen}
\begin{figure}[t]
    \centering
    \includegraphics[width=\linewidth]{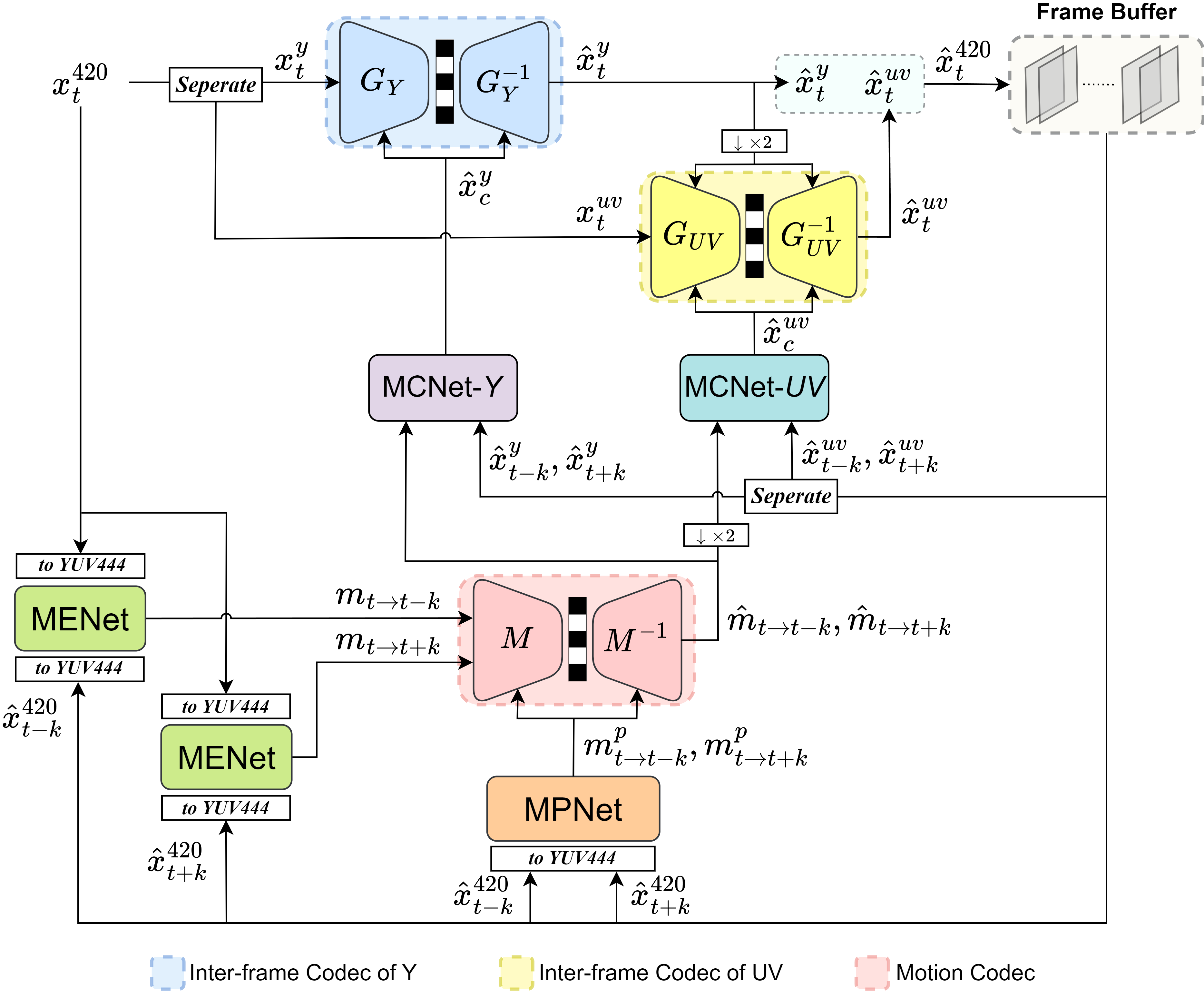}
    \caption{The proposed B-frame coding framework. $x^{420}_t$ denotes the current coding frame and $\hat{x}^{420}_{t-k},\hat{x}^{420}_{t+k}$ are the two previously reconstructed reference frames. "\textit{Separate}" is an operation that separates the Y and UV components. "$\downarrow\times2$" is the down-sampling operation by a factor of 2. Note that in down-sampling an optical flow map~($m$), 
    the values of the flow map
    are also reduced by half.}
    \label{fig:arch}
    \vspace{-1em}
\end{figure}

As illustrated in Fig.~\ref{fig:arch}, motion estimation and motion compensation are done by MENet and MCNet, respectively. In an effort to re-use PWCNet~\cite{pwc} without introducing any significant change, we interpolate the UV components and fine tune the PWCNet~\cite{pwc} to perform motion estimation in the YUV 4:4:4 domain. The flow maps thus obtained have the same resolution as the Y component and are downscaled to motion compensate the UV components. 

For motion compensation, separate MCNets are used to process Y and UV components distinctively. Similar to the motion compensation network in~\cite{dvclu}, 
\textcolor{black}{our MCNet includes as inputs the Y/UV components from the future and past reference frames, and the decoded flow maps, which are used for bi-directional backward warping.} 
Notably, we reduce the number of channels from 64 to 48 to avoid an excessive increase in model size due to the use of two MCNets.

    

\subsection{Conditional Inter-frame Coding in the YUV Domain}
\label{sec:inter_coding}
Two conditional inter-frame codecs are used to code the Y and UV components separately. \textcolor{black}{They follow the CANF-based coding schemes in Figs.~\ref{fig:canfvc} (a) and (b), respectively, where $x_c$ denotes the motion-compensated Y or UV components.} Because the Y component preserves most of the information, we use the decoded Y component $\hat{x}^y_t$ as an additional conditioning signal $x_a$ in Fig.~\ref{fig:canfvc} (b) when coding the UV components. \textcolor{black}{Notably, we remove the hyperprior~\cite{minnen2018} from the UV codec and use only the temporal prior~\cite{rlvc, nvc} to reduce complexity. A side experiment shows that this design choice has little impact (\textless 1\% rate inflation) on coding performance. }




\subsection{Adaptive Feature Modulation}
\label{sec:adaptive_module}
\begin{figure}[t!]
    \centering
    \includegraphics[width=0.75\linewidth]{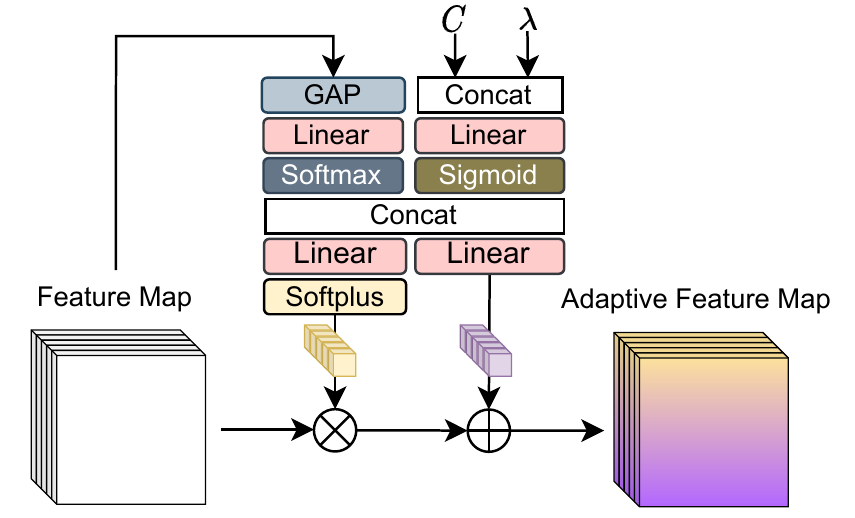}
    \caption{Adaptive feature (AF) modulation for content-adaptive variable-rate coding. $C, \lambda$ denote the coding level and the rate parameter, respectively. $\otimes,\oplus$ are the channel-wise multiplication and addition, respectively.}
    \label{fig:rate-ad-net}
\end{figure}

Adaptive feature (AF) modulation is to adapt the feature distribution in every convolutional layer, in order to achieve variable-rate compression with a single model and content-adaptive coding.
The AF modulation is placed after every convolutional layer in the motion and inter-frame codecs. As shown in Fig.~\ref{fig:rate-ad-net}, it outputs channel-wise affine parameters, which are used to dynamically adjust the output feature distributions. 

As compared to the previous works~\cite{icip21, iscas22}, our scheme has two distinctive features. One is that we introduce the coding level $C$ of a B-frame as its contextual information to achieve hierarchical rate control. This is motivated by the fact that with hierarchical B-frame coding, the reference quality of a B-frame varies with its coding level. The additional contextual information from the coding level allows greater flexibility in adjusting the bit allocation among B-frames. We note that most previous works use only a single rate parameter $\lambda$ as the contextual information without distinguishing between B-frames of different coding levels. Additionally, our AF module incorporates a global average pooling (GAP) layer to summarize the input feature maps with a 1-D feature vector. As such, our AF module is able to adapt the feature distribution in a content-adaptive manner. 


In our current implementation, $C$=0,1 has only two values because during training, there are only a limited number of coding levels (see Sec.~\ref{sec:details}). $C$=0 means the current coding B-frame will serve as a reference frame for the other B-frames at higher coding levels, while $C$=1 indicates that it is at the highest coding level and will not be utilized for reference. \textcolor{black}{See Table~\ref{tab:codingt} for an example.} Moreover, we choose $\lambda \in \{16384, 4096, 1024, 256, 128\}$ to encode B-frames at a finite number of rate points. To achieve fine-grained rate control, we incorporate an intra codec that supports continuous-step rate adaptation. To encode a video sequence at a specific rate point, we first choose from a set of pre-determined combinations of $\lambda$'s for the intra and inter codecs the one which yields a rate point matching closely the target rate. We then fine tune the $\lambda$ of the intra codec to fit the target rate precisely.

\subsection{Training Loss}
\label{sec:details}
Our training objective function is given by
\begin{equation}
\label{equ:rd_loss_P}
 L = \frac{1}{5}\sum_t{\lambda \cdot [6d(x_t^y, \hat{x}_t^y) + 2d(x_t^{uv}, \hat{x}_t^{uv})] / 8 + R_t},
\end{equation}
\textcolor{black}{where $t$ is the frame index, $d(\cdot,\cdot)$ measures the mean-squared error between the input and the output Y/UV components (the weighting factors 6 and 2 follow the evaluation metrics of the challenge~\cite{iscas_gc23}), and $R_t$ is the bit rate consumed by all the codecs. We train our scheme on 5-frame training sequences, each of which is encoded as two-level hierarchical B-frames with an intra-period 4 (see Table~\ref{tab:codingt}).}

\begin{table}[!t]
    \centering
    \caption{5-frame hierarchical B-prediction}
    \begin{tabular}{|c|c|c|c|c|c|}
         \hline
         t          & 1 & 2 & 3 & 4 & 5 \\
         \hline
         frame-type      & I & B2 ($C$=1) & B1 ($C$=0) & B2 ($C$=1) & I \\
         \hline
    \end{tabular}
    \vspace{-1em}
    \label{tab:codingt}
\end{table}

\section{Experiments}
\begin{figure*}[t!]
    \begin{center}
    \begin{subfigure}{0.24\linewidth}
        \centering
        \includegraphics[width=\linewidth]{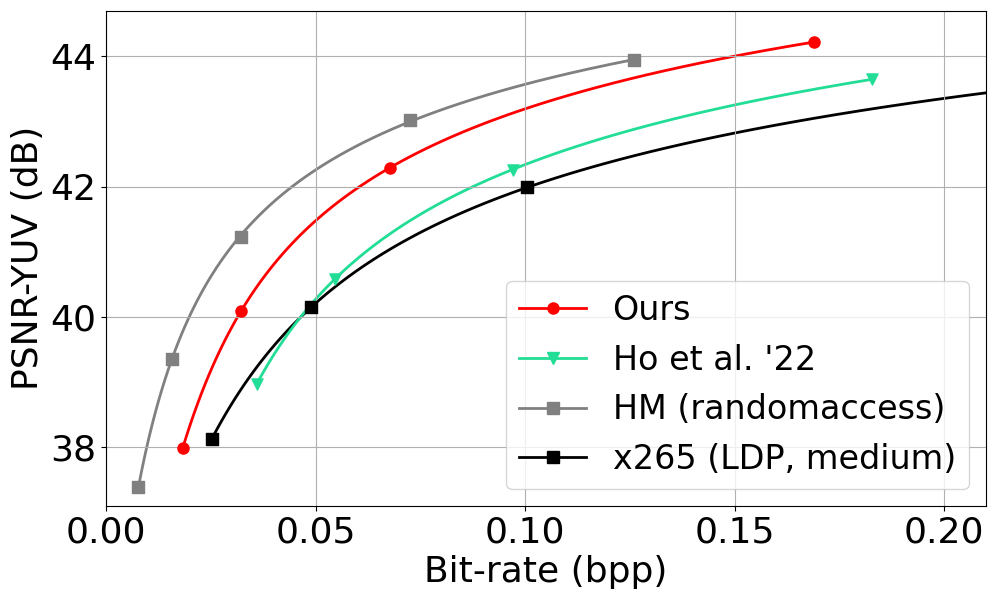}
        \caption{UVG}
        \label{fig:uvg}
    \end{subfigure}
    \begin{subfigure}{0.24\linewidth}
        \centering
        \includegraphics[width=\linewidth]{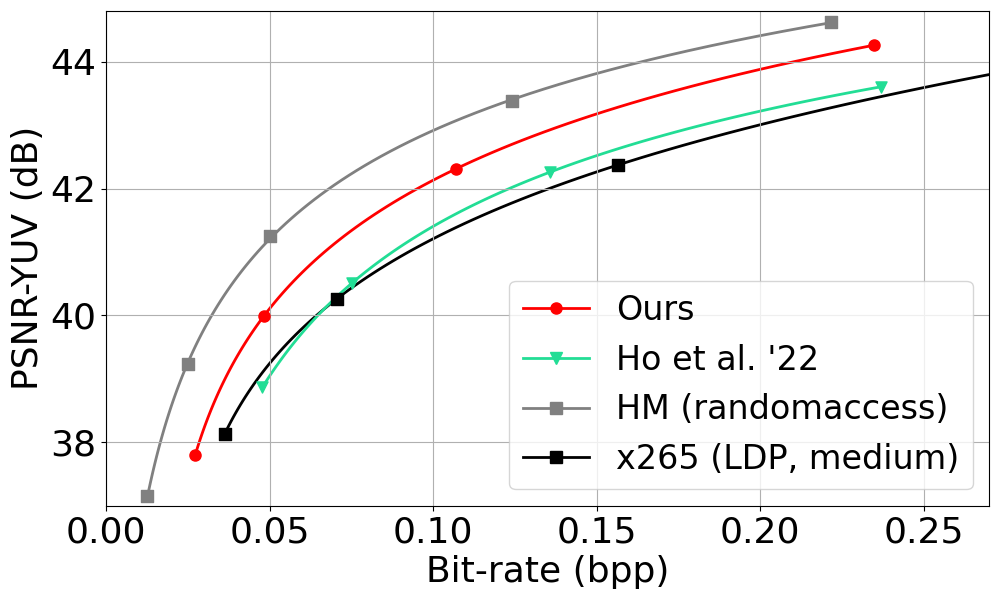}
        \caption{MCL-JCV}
        \label{fig:mcl}
    \end{subfigure}
    \begin{subfigure}{0.24\linewidth}
        \centering
        \includegraphics[width=\linewidth]{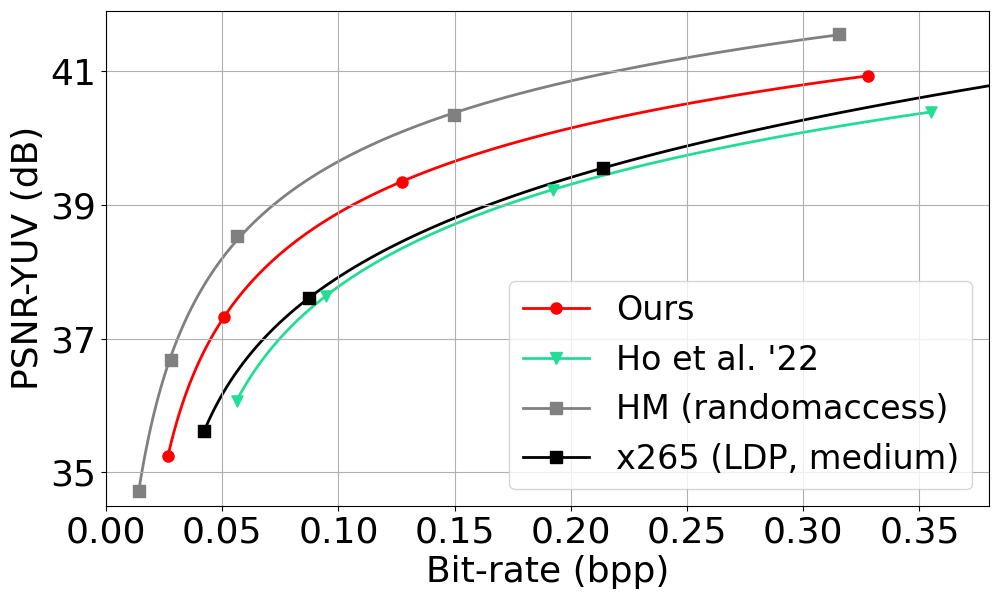}
        \caption{HEVC Class B}
        \label{fig:hevcb}
    \end{subfigure}
    \begin{subfigure}{0.24\linewidth}
        \centering
        \includegraphics[width=\linewidth]{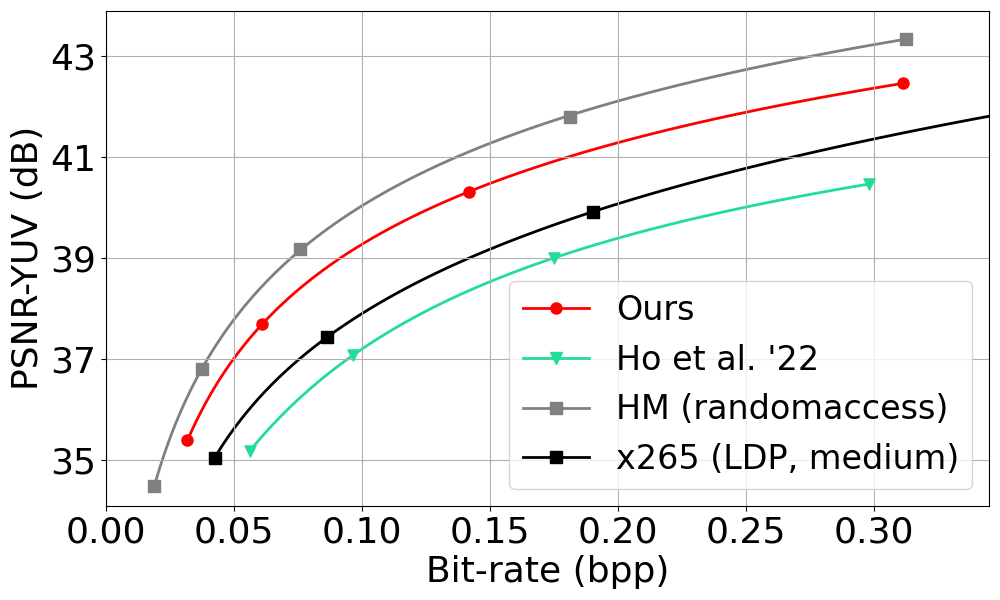}
        \caption{ISCAS'22 GC}
        \label{fig:iscas}
    \end{subfigure}
    \caption{Rate-distortion plots on UVG, MCL-JCV, HEVC Class B, and ISCAS'22 GC datasets in terms of PSNR-YUV.}
    \label{fig:RD}
    \end{center}
    \vspace{-1.05em}
\end{figure*}
\subsection{Setup}
\label{sec:implement}
We train our model on Vimeo-90k~\cite{vimeo}, where the video sequences are of size 448x256 in RGB format. For training, we randomly crop video frames to 256x256 and convert their color space from RGB to YUV 4:2:0. We use Adam optimizer~\cite{adam}, with the learning rate set to $1e^{-4}$.

For intra coding, we adopt a similar intra codec to~\cite{anfic_arxiv}. We, however, remove the context model and replace the Gaussian Mixture model with a simple Gaussian to reduce the coding runtime. 

The evaluation of compression performance is done on UVG~\cite{uvg}, MCL-JVC~\cite{mcl}, HEVC Class B~\cite{hevcctc} and ISCAS'22 Grand Challenge (GC)~\cite{iscas_gc}. All the test sequences are in YUV 4:2:0 format. The intra-period defaults to 32. We follow~\cite{iscas_gc} to calculate the PSNR, with the bit rate measured in bits per pixel (bpp).

The proposed method is compared against x265~\cite{x265} under \textit{medium} preset and \textit{low delay} configuration, and HM~\cite{HM} with \textit{encoder\_randomaccess\_main} configuration. The former is used as anchor in reporting the BD-rate numbers, unless otherwise specified. Additionally, the learned baseline method is~\cite{iscas22}, which is the top performer in ISCAS 2022 challenge~\cite{iscas_gc}. Note that~\cite{iscas22} supports P-frame coding only. We remark that none of the existing learned B-frame codecs supports YUV 4:2:0 content. 

\subsection{Experimental Results}
Fig.~\ref{fig:RD} shows the rate-distortion comparison, with the BD-rate numbers reported in Table~\ref{tab:exp_BD_PSNR}. Two observations are immediate. First, our method outperforms x265 and the learned codec in~\cite{iscas22} by a large margin across all the datasets. This is attributed to the use of more efficient B-frame coding. Second, the proposed method is seen to be inferior to HM under the random access configuration, which represents a much stronger baseline method for B-frame coding. 
\textcolor{black}{
In contrast to our comparison, the recent work~\cite{murat_lhbdc} compares with HM randomaccess in terms of PSNR-RGB with a short intra-period of 8.
We note that their experimental settings work favorably to learned codecs. To evaluate the decoded picture quality in the RGB domain, the pipeline of learned compression typically involves the color space conversion from YUV 4:2:0 as the input format to RGB for encoding and decoding. This is to be compared with the setting of HM, namely, encoding and decoding in YUV 4:2:0, followed by the color space conversion to RGB as the output format. As such, learned codecs are able to leverage end-to-end training to maximize the decoded picture quality in RGB. To compare fairly with HM, the ISCAS 2023 challenge requires that the quality evaluation be done in YUV 4:2:0.}

\begin{figure}[t]
    \centering
    \includegraphics[width=0.85\linewidth]{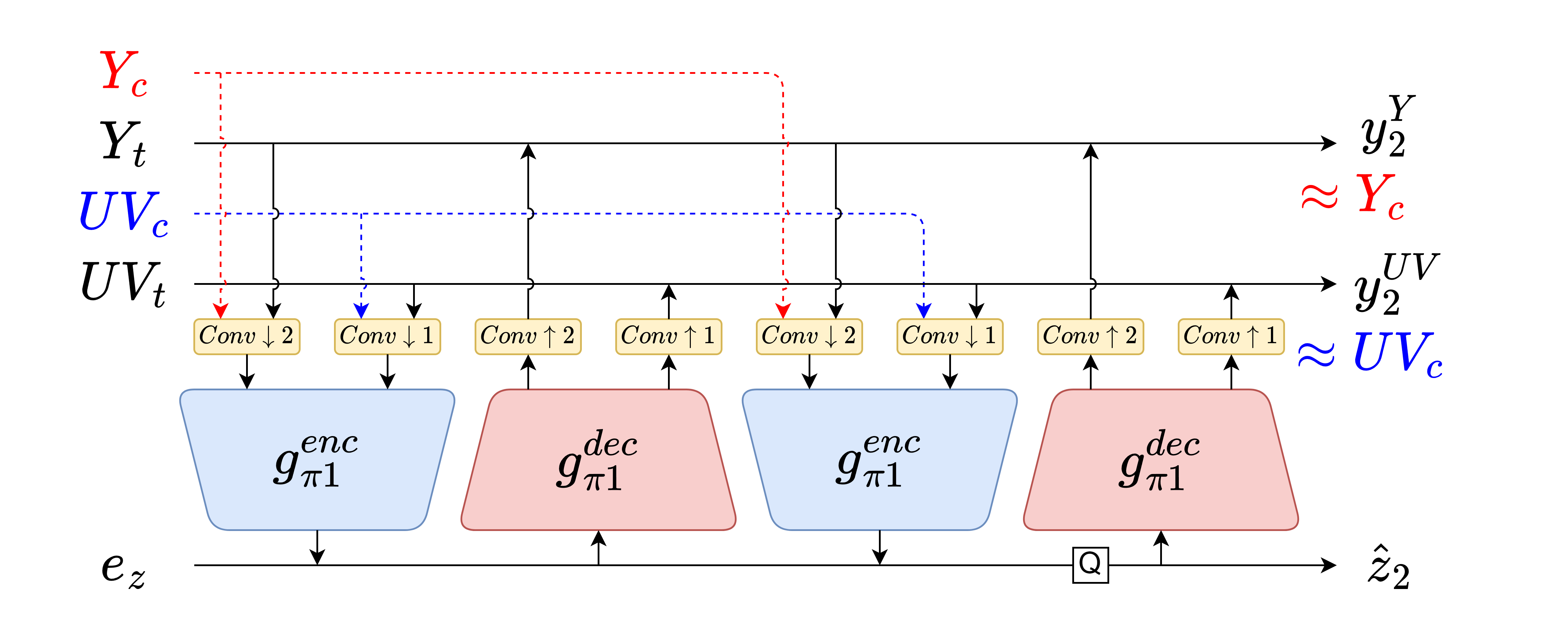}
    \caption{ \textcolor{black}{Illustration of the merged YUV coding. For brevity, the hyperprior and the temporal prior are omitted from the figure.}}
    \label{fig:merge}
    \vspace{-1.25em}
\end{figure}

Table~\ref{tab:abl_inter} presents the BD-rate comparison between our conditional YUV coding scheme and its variants, including (1) \textit{independent} coding of the Y and UV components with two separate conditional inter-frame codecs,
(2) \textit{merged} coding of the Y and UV components by converting them into their latent representations using convolutional layers for concatenation and joint coding with one conditional inter-frame codec (Fig.~\ref{fig:merge}), 
(3) \textit{space-to-depth} conversion of the Y component for concatenation with the UV components and their joint coding with one conditional inter-frame codec, and
(4) \textit{YUV 4:4:4} conversion from YUV 4:2:0 for joint coding of the Y and UV components with one conditional inter-frame codec.
For fair comparison, these competing methods have a similar model size.
\textcolor{black}{The BD-rate numbers are reported for the Y, U, and V components separately. We observe that our method outperforms all the other variants across different color components, except for the independent case, where the Y component shows slightly more rate saving (2.3\%) at the cost of much worse compression performance on the U and V components. The gain of the Y component is related to the bit allocation among these color components. Generally, the independent case suffers from much worse compression performance on the UV components. 
}


\textcolor{black}{Table~\ref{tab:abl_cl} presents an ablation study of our adaptive feature modulation, with the aim of understanding the gain from the two pieces of contextual information, i.e. the content feature and the coding level $C$ of the target frame (see Fig.~\ref{fig:rate-ad-net}). It is seen that disabling either of them or both leads to considerable rate inflation.} 

\begin{table}[!t]
    \centering
    \caption{BD-rate comparison. The anchor is x265 in LDP medium mode.}
    \label{tab:exp_BD_PSNR}
    \centerline{
    \begin{tabular}{ccccc}
        \toprule
        \multirow{2}{*}{Method}  & \multicolumn{4}{c}{BD-rate (\%) PSNR} \\
            \cline{2-5}
            & UVG   & MCL-JCV  & HEVC-B  & ISCAS'22 GC \\ 
                \hline
        Ours                            & -36.6 & -26.6 & -32.5 & -33.4 \\
        Ho~\emph{et al.}'22~\cite{iscas22} & -10.9 & -4.5  & 7.7   &  24.9 \\
        HM~\cite{HM}                    & -57.0 & -48.7 & -53.8 & -48.2 \\
        \bottomrule
    \end{tabular}
    }
\end{table}
\begin{table}[!t]
    \centering
    \caption{Ablation study on YUV coding. The dataset is ISCAS'22 GC. The anchor is our proposed method.}
    \label{tab:abl_inter}
    \centerline{
    \begin{tabular}{cccc}
        \toprule
        \multirow{2}{*}{YUV Coding} & \multicolumn{3}{c}{BD-rate (\%) PSNR} \\
            \cline{2-4}
             & Y & U & V \\ 
            \hline
            Ours             &  0.0 &  0.0 &  0.0\\
            Independent      & -2.3 &  4.0 & 17.5\\
            Merged           & 15.1 & 81.3 & 82.3\\
            Space-to-Depth   & 10.3 & 96.2 & 51.0\\
            YUV444           &  4.8 & 57.4 & 53.9\\
        \bottomrule
    \end{tabular}
    }
\end{table}

\begingroup
\setlength{\tabcolsep}{5.5pt}
\begin{table}[!t]
    \centering
    \caption{Ablation study on adaptive feature modulation. The anchor is our method with fully functional modulation.}
    \label{tab:abl_cl}
    \centerline{
    \begin{tabular}{cccccc}
        \toprule
        Content & Coding & \multicolumn{4}{c}{BD-rate (\%) PSNR} \\
            \cline{3-6}
           Adaptive & Level & UVG   & MCL-JCV  & HEVC-B  & ISCAS'22 GC \\ 
                \hline
          \checkmark  & \checkmark  & 0.0  & 0.0  & 0.0  & 0.0   \\
          \checkmark  &             & 5.4  & 5.0  & 5.1  & 8.2   \\
                      & \checkmark  & 12.3 & 9.6  & 10.0 & 10.2  \\
                      &             & 16.0 & 13.3 & 13.9 & 17.4  \\
        \bottomrule
    \end{tabular}
    \vspace{-1em}
    }
\end{table}
\endgroup

\vspace{-0.5em}
\section*{Conclusion}
We propose a hierarchical B-frame coding scheme with adaptive feature module for YUV 4:2:0 content. 
Our major findings include: (1) separately coding the Y and UV components is beneficial, (2) adapting the feature distributions to the content information and the coding levels of B-frames is crucial to content-adaptive variable-rate coding, and (3) in terms of coding YUV 4:2:0 content, our learned codec still has ample room for further improvement as compared to HM. 
\newpage
{\small
\bibliographystyle{IEEEtran}
\bibliography{egbib}
}

\end{document}